\documentclass[conference]{IEEEtran}

\usepackage{amsmath,amssymb}
\usepackage{graphicx}
\usepackage[hidelinks]{hyperref}
\usepackage{geometry}
\title{Latent Structural Similarity Networks for Unsupervised Discovery in Multivariate Time Series}

\author{
\IEEEauthorblockN{Olusegun Owoeye}
\IEEEauthorblockA{
Independent Researcher\\
University of Cambridge (current affiliation)\\
olusegun.owoeye@outlook.com
}}

\date{}

\begin{document}
\sloppy

\maketitle

\begin{abstract}
This paper proposes a task-agnostic discovery layer for multivariate time series that constructs a relational hypothesis graph over entities without assuming linearity, stationarity, or a downstream objective. The method learns window-level sequence representations using an unsupervised sequence-to-sequence autoencoder, aggregates these representations into entity-level embeddings, and induces a sparse similarity network by thresholding a latent-space similarity measure. This network is intended as an analyzable abstraction that compresses the pairwise search space and exposes candidate relationships for further investigation, rather than as a model optimized for prediction, trading, or any decision rule. The framework is demonstrated on a challenging real-world dataset of hourly cryptocurrency returns, illustrating how latent similarity induces coherent network structure; a classical econometric relation is also reported as an external diagnostic lens to contextualize discovered edges.
\end{abstract}


\section{Introduction}

Many scientific and industrial systems generate high-dimensional multivariate time series whose dynamics evolve across entities and operating conditions.
Examples include financial assets, biological measurements, engineered components subject to degradation, and sensor-equipped industrial assets.
A central challenge in such settings is unsupervised relational discovery: identifying which entities exhibit similar temporal behaviour when the data are non-stationary and strong modelling assumptions are undesirable.

Relational discovery is often approached by applying fixed similarity measures directly in the observation space.
In financial applications, this frequently corresponds to univariate return-based representations that simplify modelling and facilitate large-scale analysis \cite{deepFinancialNetwork2024}.
Such formulations implicitly assume that returns alone capture the relationships of interest.
However, reducing multivariate behaviour to a single series can discard informative temporal and cross-sectional structure, including intra-window dynamics and volatility patterns, that may distinguish entities with superficially similar average co-movement.

In financial markets, a canonical instance of relational discovery is pair screening, where the aim is to identify asset pairs exhibiting coherent relative behaviour.
Classical approaches rely on distance measures, correlation, or cointegration tests \cite{gatev2006pairs,engle1987cointegration}.
While effective in controlled settings, these methods can be sensitive to window selection and regime shifts, are tied to linear dependence assumptions, and become cumbersome as the entity universe grows.
These properties make financial markets a demanding environment for stress-testing general discovery pipelines.

Representation learning offers an alternative route by transforming raw sequences into low-dimensional embeddings that can capture non-linear temporal structure.
Unsupervised models such as autoencoders and recurrent architectures provide flexible mechanisms for mapping complex dynamics into latent spaces \cite{hinton2006reducing}.
Such embeddings have been used for dimensionality reduction, anomaly detection, and regime characterization in time series, and have increasingly been used as inputs to descriptive network analyses in finance.

Despite this progress, representation learning and relational analysis are often treated as separate ends: a latent space is learned, and relational structure is then examined as an auxiliary product or for a downstream task.
Less attention has been paid to using learned representations explicitly as a discovery layer, where the primary object of interest is the induced relational structure itself, examined independently of prediction, optimisation, or decision rules.
In particular, the role of sequence embeddings in supporting entity-level discovery under non-stationarity remains underdeveloped.

This paper proposes an unsupervised discovery framework that integrates sequence-based representation learning with similarity network construction to uncover latent relational structure in multivariate time series.
Rolling windows are used to form sequence inputs that accommodate non-stationarity.
Window-level sequences are embedded into a shared latent space via a sequence-to-sequence Long Short-Term Memory (LSTM) autoencoder, and entity-level embeddings are formed by aggregating window representations.
Pairwise similarity in latent space is then used to induce a sparse similarity graph, providing an analyzable relational hypothesis layer that compresses the dense pairwise search space into a tractable set of candidate relationships.
The framework is intentionally task-agnostic: it is not designed to maximise predictive accuracy, profitability, or any other downstream objective, and it does not claim that the induced graph is a ground-truth structure.

Hourly cryptocurrency market data are used as a demonstration domain due to their scale, noise, and pronounced non-stationarity, as well as the existence of well-studied relational concepts in the literature.
The demonstration focuses on the topology of the induced similarity networks and on post-discovery statistical diagnostics used only as contextual lenses.
While finance provides a demanding benchmark for implementation, the methodology is domain-agnostic and directly applicable to settings such as biological phenotyping, materials degradation analysis, and industrial asset monitoring.

\section{Background and Related Work}

The unsupervised discovery of structural relationships in time series data has been studied across a wide range of domains, including finance, engineering systems, and the natural sciences.
A common objective in these settings is to identify entities that exhibit related temporal behaviour, often as a precursor to further analysis or decision-making.
Traditional approaches typically rely on fixed similarity measures applied directly to observed data, with similarity defined through correlation, distance metrics, or shared stochastic trends.
In financial applications, this problem has most prominently appeared in the context of pair discovery, where correlation-based screening and cointegration analysis have served as canonical tools \cite{gatev2006pairs,engle1987cointegration}.
While these methods are well understood and computationally efficient, their performance depends strongly on modelling assumptions, window selection, and the persistence of linear relationships over time.

A central limitation of classical similarity-based approaches lies in their reliance on predefined representations.
In many financial studies, similarity is evaluated using univariate return series, which simplifies analysis but abstracts away intra-period dynamics, volatility structure, and cross-sectional interactions.
Such representations implicitly assume that the salient structure governing relational similarity is fully captured by returns alone.
Under non-stationarity or regime shifts, this assumption may be restrictive, as entities can exhibit similar higher-order dynamics despite weak contemporaneous correlation.

Representation learning offers an alternative perspective by learning data-driven embeddings that aim to capture salient temporal structure in a lower-dimensional space.
Unsupervised neural architectures, including autoencoders and recurrent models, have been widely applied to multivariate time series for dimensionality reduction, anomaly detection, and regime identification \cite{hinton2006reducing,malhotra2016lstm}.
Sequence-to-sequence autoencoders in particular provide a flexible mechanism for encoding temporal dependencies across multiple variables, allowing complex dynamics to be compressed into latent representations that can be compared across entities.

The use of learned embeddings naturally motivates similarity evaluation in latent space.
Prior work has shown that distance-based measures applied to latent representations can provide meaningful proxies for structural similarity, particularly when embeddings are trained to preserve temporal or dynamical characteristics of the underlying data \cite{franceschi2019unsupervised}.
In this setting, similarity is no longer defined directly on raw observations, but on representations that integrate information across time and features.
This shift enables relational analysis that is less sensitive to noise and more robust to superficial differences in scale or timing.

Latent representations have also been used as a basis for network construction, where entities are treated as nodes and similarities define weighted edges.
Such approaches have been explored in a range of contexts, including graph construction from embeddings for clustering, system-level analysis, and exploratory discovery \cite{hamilton2017representation}.
In financial applications, recent work has combined encoder-based models with network analysis to study large-scale structural evolution across market regimes \cite{deepFinancialNetwork2024}.
These studies demonstrate the utility of deep representations for constructing interpretable relational structures, but they typically focus on descriptive analysis or compression rather than on the stability and persistence of discovered relationships.

Across both classical and deep learning-based approaches, relatively limited attention has been given to representation learning as an explicit discovery layer.
Most existing work either fixes the representation and analyses relationships, or learns embeddings without explicitly evaluating the consistency of the induced relational structure under rolling or time-varying conditions.
As a result, research on the suitability of learned representations for sustained unsupervised discovery remains scarce.

The framework proposed in this paper is motivated by this unadressed area.
Rather than treating representation learning and relational analysis as separate objectives, the approach integrates sequence-based embedding with network construction to support unsupervised entity-level discovery.
By emphasising rolling evaluation and structural coherence, the framework isolates the discovery problem from downstream optimisation or decision-making.
Although financial data provide a challenging and well-studied testbed, the methodological contribution is domain-agnostic and directly applicable to settings such as biological phenotyping, materials lifetime degradation analysis, and industrial asset monitoring.

\section{Methodology}

This section describes the proposed framework for unsupervised relational discovery in multivariate time series.
The methodology integrates sequence-based representation learning with similarity network construction to identify structurally similar entities.
The framework is intentionally designed to isolate the discovery problem from downstream optimisation or decision-making, so that the coherence and stability of the discovered relational structure can be examined independently of any predictive or trading objective.

\subsection{Problem Setup}

Consider a collection of $N$ entities, each associated with a multivariate time series.
For entity $i$, observations are given by $\mathbf{X}_i = \{\mathbf{x}_{i,t}\}_{t=1}^{T}$, where $\mathbf{x}_{i,t} \in \mathbb{R}^d$ denotes a $d$-dimensional feature vector observed at time $t$.
The objective is to identify pairs or groups of entities that exhibit similar temporal structure without supervision and without imposing strong prior assumptions on the form of similarity.

Rather than defining similarity directly on raw observations, the framework seeks to discover relationships in a learned latent space that captures relevant temporal dynamics.
Discovery is performed using representations aggregated across rolling windows to support interpretable relational analysis.

\subsection{Windowing and Normalisation}

To support representation learning under non-stationarity, each entity’s time series is segmented into overlapping windows of fixed length $L$.
Windowing increases the effective sample size and enables local temporal structure to be captured without assuming global stationarity.
For a given window indexed by $\tau$, the corresponding segment is denoted $\mathbf{X}_i^{(\tau)} \in \mathbb{R}^{L \times d}$.

Within each window, features are normalised independently to remove scale effects and facilitate comparison across entities.
This ensures that similarity reflects structural dynamics rather than absolute magnitudes and improves robustness to regime shifts across time and entities.

\subsection{Sequence-to-Sequence Representation Learning}

Each windowed multivariate sequence is embedded into a low-dimensional latent representation using a sequence-to-sequence LSTM autoencoder.
The encoder maps the input sequence to a latent vector $\mathbf{h}_i^{(\tau)} \in \mathbb{R}^k$, where $k \ll Ld$, and a corresponding decoder is trained to reconstruct the original sequence.

An LSTM-based architecture is selected because it explicitly models temporal dependencies and non-linear dynamics that are not captured by linear dimensionality reduction techniques or static encoders.
This makes it well suited to multivariate time series in which structural similarity depends on the evolution of patterns over time rather than instantaneous feature correlations.
The autoencoder is trained in an unsupervised manner by minimising a reconstruction loss across all entities and windows, enforcing compression while preserving temporal structure.

\subsection{Latent Aggregation and Similarity}

For each entity, latent embeddings $\mathbf{h}_i^{(\tau)}$ obtained across rolling windows are aggregated to produce a single representative vector
\begin{equation}
\mathbf{z}_i = \frac{1}{K_i} \sum_{\tau=1}^{K_i} \mathbf{h}_i^{(\tau)},
\end{equation}
where $K_i$ is the number of windows associated with entity $i$.
Aggregation reduces sensitivity to transient local variation and yields stable per-entity representations suitable for similarity analysis.

Pairwise similarity between entities is computed in latent space using cosine similarity,
\begin{equation}
s_{ij} = \frac{\mathbf{z}_i^\top \mathbf{z}_j}{\|\mathbf{z}_i\|\,\|\mathbf{z}_j\|}.
\end{equation}
Cosine similarity emphasises directional alignment and proportional co-movement in latent dynamics while remaining invariant to absolute magnitude differences, aligning similarity assessment with relative structural behaviour rather than scale-dependent variation.

\subsection{Similarity Network Construction}

A weighted, undirected graph is constructed by treating entities as nodes and latent similarities as weighted edges.
Representing similarity structure as a graph provides a compact and interpretable abstraction of inter-entity relationships and enables the use of generic graph-based analysis tools.

To obtain an interpretable and non-degenerate network, a similarity threshold $\tau$ is selected empirically to avoid graphs that are overly sparse or overly dense.
In the reported experiments, a single representative threshold is fixed for consistency and comparability across reported results; alternative sparsification schemes and threshold sweeps are natural extensions.

The resulting network provides a structured representation of inter-entity relationships.
Downstream analysis may involve clustering, community detection, or other graph-based techniques selected according to the requirements of the application domain.
This paper focuses on the construction and analysis of the similarity network itself rather than on any specific downstream grouping or decision procedure.

\subsection{Robustness and Stability Protocol}

Since the framework is intended for discovery rather than optimisation, robustness is framed in terms of the stability of the induced relational structure under reasonable perturbations.
The checks below define a protocol for assessing stability without reference to any downstream objective.

\textbf{Temporal stability.}
The similarity network can be reconstructed over multiple contiguous time blocks within the study period using the same windowing, normalisation, and representation learning procedure.
Stability can then be quantified by measuring agreement between graphs across blocks, for example via edge-set overlap (e.g., Jaccard similarity) evaluated at a fixed sparsity level, or via rank correlation of similarity matrices.
Here, a \emph{core set of edges} refers to edges that persist across multiple re-estimations (e.g., appearing in a majority of time blocks under a fixed sparsification rule).

\textbf{Sensitivity to thresholding.}
Since sparsification is performed by thresholding latent cosine similarities, robustness can be examined by varying $\tau$ within a high-similarity regime and checking whether macroscopic graph properties are preserved.
Examples include the size of the largest connected component, the presence of dominant hubs, and the persistence of high-confidence edges.
This assesses whether qualitative relational structure is overly dependent on a single threshold choice.

\textbf{External diagnostic lens.}
Finally, discovered edges may be compared against relationships identified by classical econometric diagnostics such as Engle--Granger cointegration.
Agreement provides a sanity check that the learned representation recovers some known linear equilibrium relationships, while the framework remains agnostic to the specific form of dependence encoded in the latent space.

\subsection{Scope and Design Boundaries}

The proposed methodology is intentionally limited to unsupervised representation learning and relational discovery.
No assumptions are made regarding downstream optimisation, decision rules, or execution strategies.
Accordingly, evaluation emphasises the coherence and stability of the induced similarity structure rather than task-specific predictive performance.
This design choice allows the discovery mechanism to be examined independently and facilitates application across domains where the interpretation and use of discovered relationships may differ.

\section{Experiments}

This section demonstrates the proposed latent representation and similarity network framework in a controlled financial setting.
The objective is to characterise the relational structure induced by the discovery layer on a challenging multivariate time series dataset, without incorporating downstream optimisation, execution, or trading logic.

\subsection{Data Universe}

The experimental universe consists of cryptocurrency spot market time series data.
Assets are selected as the top 20 cryptocurrencies by market capitalisation over the study period, quoted against a USDT quote currency.
All data is sourced from the Binance exchange to ensure sufficient liquidity and data continuity.

The observation window spans one year, from April 2024 to April 2025.
All experiments reported in this paper are conducted exclusively at the 1-hour sampling frequency.
This resolution is selected to balance temporal granularity with statistical robustness and to provide a controlled setting for demonstrating the discovery mechanism.
Extension to alternative sampling frequencies is left for future work.

\subsection{Preprocessing and Window Construction}

For each asset, raw OHLC price series are transformed into a stationary representation using logarithmic returns computed from Open, High, Low, and Close prices.
No additional feature channels are introduced.
The resulting multivariate time series is segmented into overlapping rolling windows of fixed length $L = 30$ timesteps, consistent with the methodology described in Section~III.

Each rolling window is treated as an independent input sample for representation learning.

\subsection{Latent Representation Learning}

A sequence-to-sequence LSTM autoencoder is trained in an unsupervised manner using rolling windows pooled from the full asset universe.
A single shared model is used for all assets at the 1-hour timeframe, producing a common latent space in which cross-asset comparisons are meaningful.

For each rolling window, the encoder outputs a latent embedding that compresses the temporal structure of the input sequence.
Latent embeddings are then aggregated across windows to obtain a single representative latent vector per asset.
This aggregation step reduces sensitivity to transient local variation and yields stable per-asset representations for similarity analysis.

\textbf{Implementation details.}
The autoencoder is implemented in TensorFlow/Keras and trained to minimise mean squared reconstruction error using the Adam optimiser.
The encoder comprises two stacked LSTM layers with hidden size 256; the final LSTM outputs a fixed-length vector that is projected to a 64-dimensional latent representation via a ReLU-activated dense layer.
The decoder maps the latent vector back to sequence space using a dense projection and repetition over the sequence length, followed by two stacked LSTM layers and a time-distributed linear layer to reconstruct the original feature dimension.
Training uses mini-batches of size 64 for 20 epochs on the pooled rolling-window dataset to produce a shared latent space for cross-entity comparison.

The purpose of this stage is dimensionality reduction and structural feature extraction rather than prediction.

\subsection{Similarity Network Construction}

Pairwise similarity between assets is computed using cosine similarity applied to the aggregated latent representations.
A weighted, undirected similarity graph is constructed by treating assets as nodes and latent similarities as weighted edges.

To obtain an analyzable graph, edges are retained using a fixed similarity threshold selected empirically to avoid networks that are either excessively sparse or overly dense.
The threshold is not optimised against any downstream objective as part of the experimental demonstration.
Candidate relationships are extracted directly from the resulting graph structure.

\subsection{Post-Discovery Statistical Diagnostics}

As a contextual diagnostic lens, an Engle--Granger cointegration test is applied \emph{after} the network-based discovery stage to candidate pairs surfaced by the similarity graph.
The diagnostic outcome is reported descriptively as the number (and proportion) of discovered candidate pairs that satisfy the cointegration criterion under a fixed testing protocol.
This diagnostic is not treated as a target, optimisation objective, or ground-truth label.
Cointegration is considered neither a necessary nor a sufficient condition for latent structural similarity.

\subsection{Experimental Scope}

The experiments are intentionally restricted to demonstrating the latent representation and similarity discovery mechanism.
No comparisons against alternative discovery pipelines are performed, and no downstream optimisation or trading performance is evaluated.
Robustness considerations are treated at the level of protocol: Section~III-F specifies temporal stability and threshold-sensitivity checks that can be applied to the discovery layer, but this version reports results at a single sampling frequency and a single fixed similarity threshold for clarity and comparability.
This design isolates the behaviour of latent structural similarity learning and supports applicability across domains beyond finance.

\section{Results}

This section presents the results obtained by applying the proposed latent similarity framework to the 1-hour cryptocurrency dataset described in Section~IV.
The results characterise the latent similarity structure, the induced network topology, and the properties of relationships surfaced by the discovery process.

\subsection{Latent Similarity Structure}

\begin{figure}[t]
    \centering
    \includegraphics[width=0.95\linewidth]{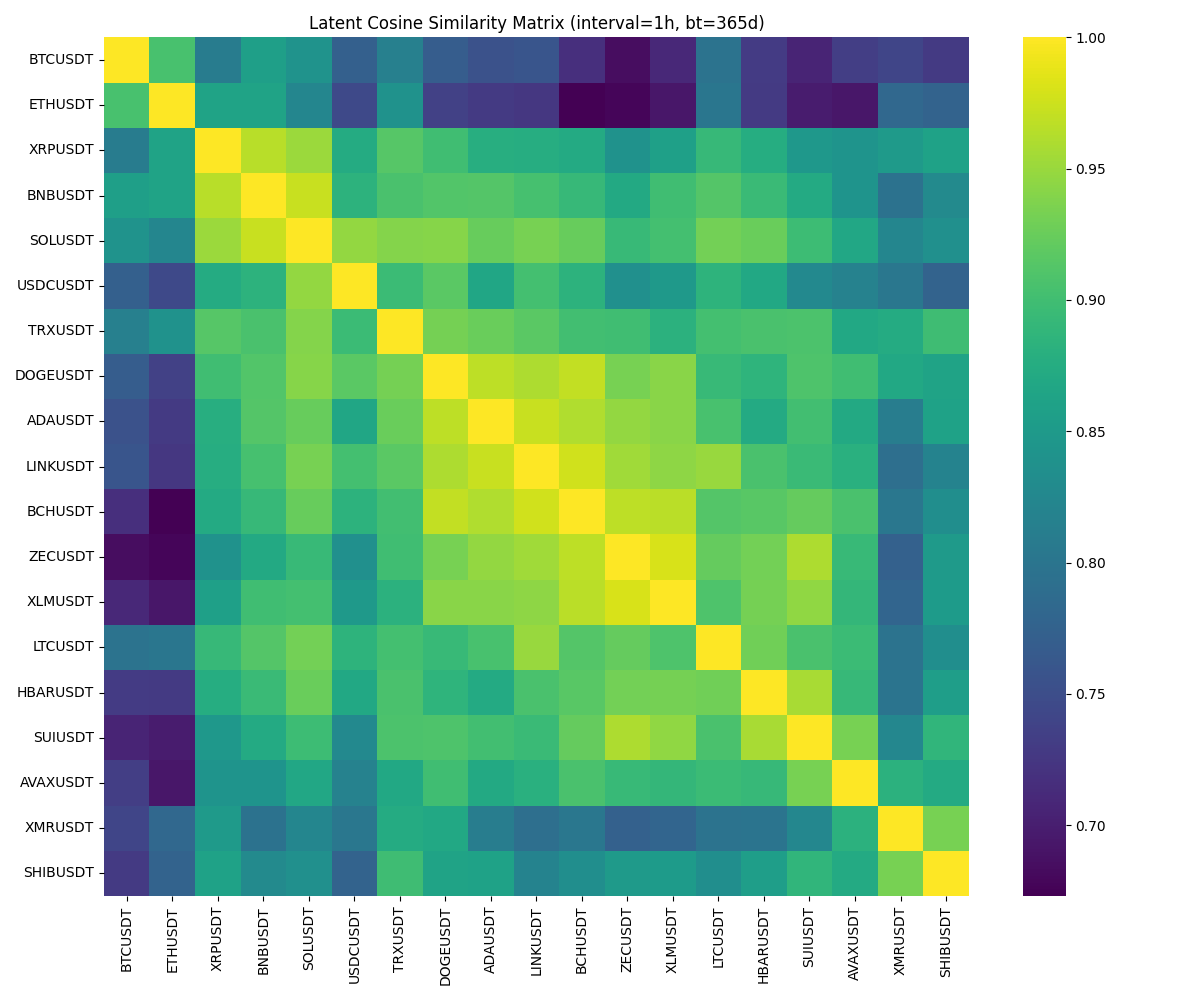}
    \caption{Cosine similarity matrix computed between aggregated latent representations for 20 cryptocurrency assets at the 1-hour resolution.}
    \label{fig:similarity_matrix}
\end{figure}

Figure~\ref{fig:similarity_matrix} shows the cosine similarity matrix computed between aggregated latent representations for the full asset universe.
Non-uniform structure is visible in the matrix, with subsets of assets exhibiting relatively high latent similarity.
This indicates that the learned latent space differentiates assets based on shared temporal characteristics rather than collapsing to a uniform market-wide representation.

The similarity distribution exhibits separation between more strongly aligned and more weakly aligned asset pairs, motivating the use of thresholding to induce a sparse and interpretable similarity network.

\subsection{Similarity Network Characteristics}

Applying a fixed cosine similarity threshold of $\tau = 0.90$ yields a sparse weighted undirected graph over the 20 assets.
The resulting network contains multiple connected components, including both dense subgraphs and isolated pairs.

A total of 64 candidate relationships (edges) are retained under this threshold.
Connectivity is heterogeneous across assets, with some assets exhibiting relatively high local connectivity while others remain weakly connected or isolated.
This non-uniform topology reflects differences in latent similarity across entities rather than uniform exposure to common market effects.

\begin{figure}[t]
    \centering
    \includegraphics[width=0.95\linewidth]{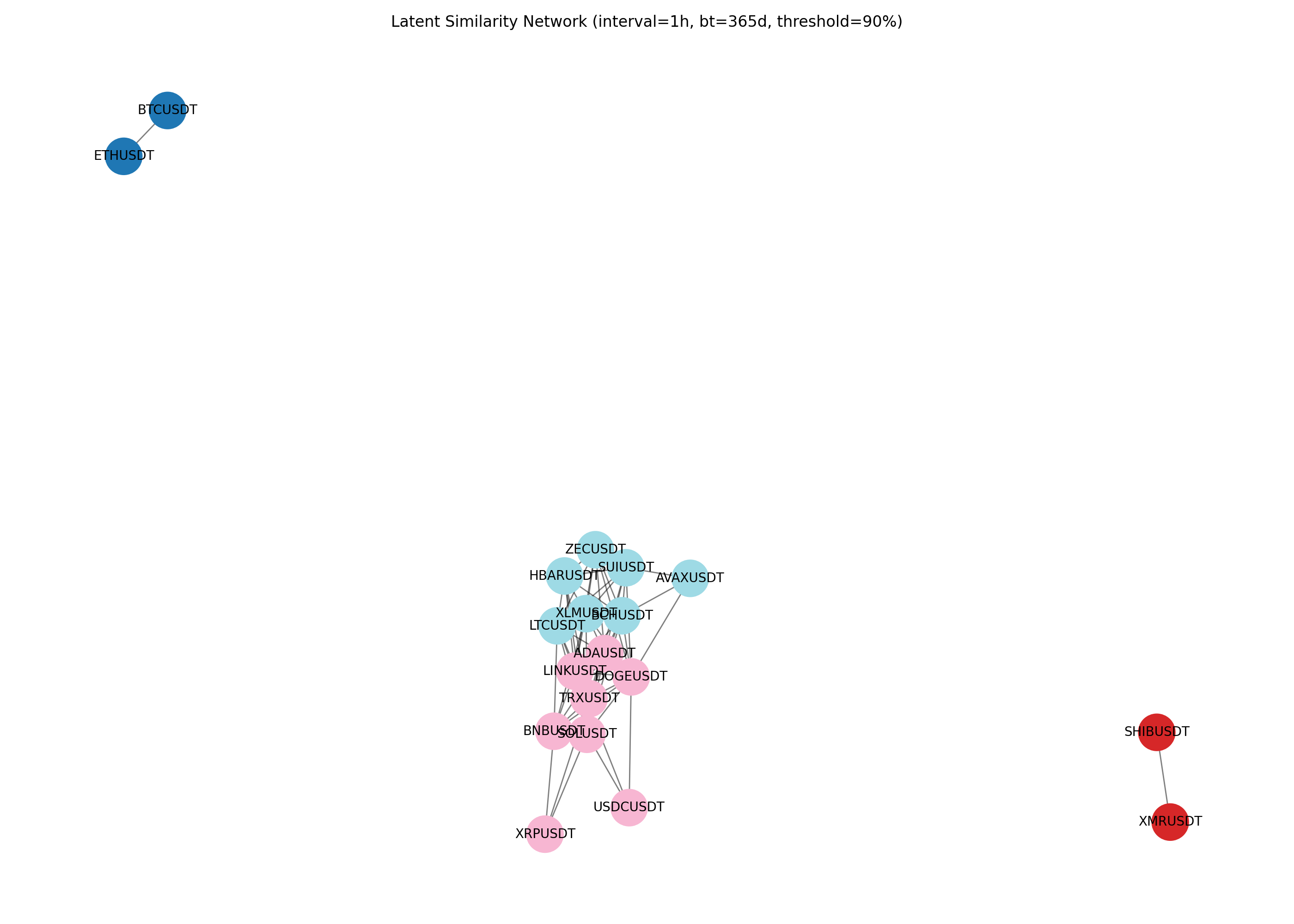}
    \caption{Latent similarity network induced using a cosine similarity threshold of 0.90. Nodes represent assets and edges denote retained latent similarity relationships.}
    \label{fig:similarity_network}
\end{figure}

The induced similarity network is illustrated in Figure~\ref{fig:similarity_network}.

\subsection{Post-Discovery Cointegration Diagnostics}

To examine how the discovered relationships relate to classical statistical notions of pair structure, cointegration testing is applied \emph{after} the network-based discovery stage.
Each of the 64 candidate pairs identified by the similarity network is evaluated using an Engle--Granger cointegration test at a fixed confidence level of $0.95$.

Out of the 64 discovered pairs, 16 satisfy the cointegration criterion.
The remaining pairs fail standard linear cointegration diagnostics despite exhibiting strong latent similarity.

This result indicates that the discovery framework surfaces a mixture of relationships, including pairs that satisfy classical linear cointegration criteria and pairs whose similarity is not captured by such tests.
Cointegration is therefore neither a necessary nor a sufficient condition for latent structural similarity and is used here strictly as a post-hoc diagnostic rather than as an optimisation objective.

\subsection{Summary}

Overall, the results show that the latent similarity framework produces a sparse and interpretable discovery layer.
By constraining analysis to a subset of relationships exhibiting strong latent alignment, the method enables focused post-discovery investigation without requiring exhaustive evaluation of all possible entity pairs.

\section{Discussion}

This section discusses the implications of the observed network structure and the role of latent similarity as a discovery mechanism.
The discussion focuses on properties of the induced similarity graph rather than on the internal latent representations themselves, which are treated as an intermediate abstraction.

\subsection{Interpretability of Network Structure}

The similarity network induced by latent cosine similarity exhibits a structured and interpretable topology.
At the chosen threshold, the graph decomposes into multiple connected components, including dense subgraphs and isolated pairs.
This heterogeneity indicates that the framework differentiates entities based on shared temporal characteristics encoded in the latent representations, rather than imposing uniform connectivity across the universe.

Importantly, the network representation provides an interpretable interface for analysing relationships.
Edges correspond to strong latent alignment, while the absence of edges explicitly restricts the space of candidate relationships.
This supports the use of network structure as a discovery layer that constrains downstream analysis without prescribing specific actions or optimisation rules.

\subsection{Relationship to Classical Statistical Diagnostics}

Post-discovery cointegration diagnostics indicate that only a subset of network-derived candidate pairs satisfy classical linear stationarity criteria.
This result highlights a distinction between latent structural similarity and traditional statistical definitions of pair structure.

Rather than interpreting non-cointegrated pairs as false positives, these findings suggest that latent similarity reflects a broader notion of shared temporal behaviour.
Cointegration is therefore best viewed as a diagnostic signal applied after discovery rather than as a target or optimisation objective.
This distinction is particularly relevant in domains where relationships may be non-linear or structurally complex.

\subsection{Role of Thresholding and Sparsity}

Thresholding plays a central role in producing interpretable similarity networks.
By avoiding excessively dense or overly sparse graphs, the framework yields structures that balance coverage with selectivity.
While the choice of threshold influences network topology, the observed separation between higher and lower similarity values suggests that the induced structure reflects non-uniform organisation in latent space rather than trivial graph artefacts.

From this perspective, the resulting network may be viewed as a parsimonious projection of latent similarity relationships rather than as a precise or exhaustive representation.

\subsection{Relation to Prior Frameworks}

The latent similarity and network construction components examined in this work were originally developed as part of a broader proprietary research framework for systematic relational discovery.
Within that framework, these components serve as an upstream abstraction layer that produces structured candidate relationships for further analysis.

The present study intentionally isolates and generalises this discovery mechanism, removing domain-specific optimisation and decision logic in order to examine its properties independently.
By abstracting the method away from its original application context, this paper demonstrates that latent structural similarity networks can function as a standalone analytical tool applicable across a range of multivariate time series settings.

\subsection{Generality Beyond the Financial Domain}

Although financial time series provide a challenging and data-rich testbed, the discovery mechanism described in this work is not domain-specific.
The framework relies solely on multivariate sequential observations and does not assume any financial interpretation of inputs or outputs.

As a result, similar network-based discovery pipelines may be applicable to other domains involving multivariate temporal processes, such as biological phenotyping, industrial asset monitoring, or materials degradation analysis.
In such settings, network structure may serve as an exploratory tool for identifying coherent behavioural groups or candidate relationships prior to domain-specific validation.

\subsection{Limitations}

The present study focuses on a single sampling frequency and a fixed similarity threshold in order to isolate the discovery mechanism.
Extensions to alternative temporal resolutions, adaptive sparsification strategies, or comparative evaluation across different similarity constructions are left for future work.
Additionally, while cointegration diagnostics provide a useful point of reference, they represent only one of many possible validation signals and should not be interpreted as exhaustive.

\section{Conclusion}

This paper presented a general unsupervised discovery framework for identifying latent relational structure in multivariate time series.
By combining sequence-based representation learning with similarity network construction, the framework provides an interpretable abstraction for relational discovery that is decoupled from downstream optimisation or decision-making.

Using a sequence-to-sequence LSTM autoencoder, multivariate temporal observations are embedded into a shared latent space, from which pairwise similarities are computed and used to induce a sparse similarity network.
Applied to financial market data as a demonstration domain, the resulting network exhibits non-uniform structure that restricts the relational search space and enables focused post-discovery analysis.
Post-hoc cointegration diagnostics illustrate that the discovered relationships include both classically stationary pairs and relationships not captured by linear statistical tests, highlighting the distinction between latent structural similarity and traditional pair definitions.

The contribution of this work lies in isolating latent similarity networks as a standalone discovery layer for multivariate time series analysis.
By separating representation learning and relational discovery from task-specific objectives, the framework provides a flexible tool that can be adapted to a range of domains where exploratory identification of structurally related entities is of interest.

While the present study focuses on a single sampling frequency and a fixed sparsification strategy, the framework itself is not restricted to financial data or to a particular similarity construction.
Future work may explore alternative representation models, sparsification schemes, or domain-specific validation criteria.
More broadly, latent similarity networks offer a principled approach for structuring complex multivariate temporal systems prior to domain-dependent modelling and decision-making.

\bibliographystyle{IEEEtran}
\bibliography{references}

\end{document}